\documentclass[10pt,twocolumn,letterpaper]{article}

\usepackage[table,xcdraw]{xcolor}
\usepackage{cvpr}
\usepackage{times}
\usepackage{epsfig}
\usepackage{graphicx}
\usepackage{amsmath}
\usepackage{amssymb}
\usepackage{enumitem}
\usepackage{mathrsfs}
\usepackage{lipsum}

\usepackage[pagebackref=true,breaklinks=true,letterpaper=true,colorlinks,bookmarks=false]{hyperref}

 \cvprfinalcopy 

\def\cvprPaperID{4020} 

\ifcvprfinal\fi
\begin{document}

\title{Train, Diagnose and Fix: Interpretable Approach for Fine-grained Action Recognition}

\author{Jingxuan Hou\thanks{Both authors contributed equally to this work.}\\
Tsinghua University\\
Beijing, China\\
{\tt\small houjx14@mails.tsinghua.edu.cn}
\and
Tae Soo Kim\footnotemark[1]\\
Johns Hopkins University\\
Baltimore, MD\\
{\tt\small tkim60@jhu.edu}
\and
Austin Reiter\\
Johns Hopkins University\\
Baltimore, MD\\
{\tt\small areiter@cs.jhu.edu}
}

\maketitle

\begin{abstract}
   Despite the growing discriminative capabilities of modern deep learning methods for recognition tasks, the inner workings of the state-of-art models still remain mostly black-boxes. In this paper, we propose a systematic interpretation of model parameters and hidden representations of Residual Temporal Convolutional Networks (Res-TCN) for action recognition in time-series data. We also propose a Feature Map Decoder as part of the interpretation analysis, which outputs a representation of model's hidden variables in the same domain as the input. Such analysis empowers us to expose model's characteristic learning patterns in an interpretable way. For example, through the diagnosis analysis, we discovered that our model has learned to achieve view-point invariance by implicitly learning to perform rotational normalization of the input to a more discriminative view. Based on the findings from the model interpretation analysis, we propose a targeted refinement technique, which can generalize to various other recognition models. The proposed work introduces a three-stage paradigm for model learning: training, interpretable diagnosis and targeted refinement. We validate our approach on skeleton based 3D human action recognition benchmark of NTU RGB+D. We show that the proposed workflow is an effective model learning strategy and the resulting Multi-stream Residual Temporal Convolutional Network (MS-Res-TCN) achieves the state-of-the-art performance on NTU RGB+D.
   
\end{abstract}

\section{Introduction}

\begin{figure}[t]
\begin{center}
 \includegraphics[width=8.5cm]{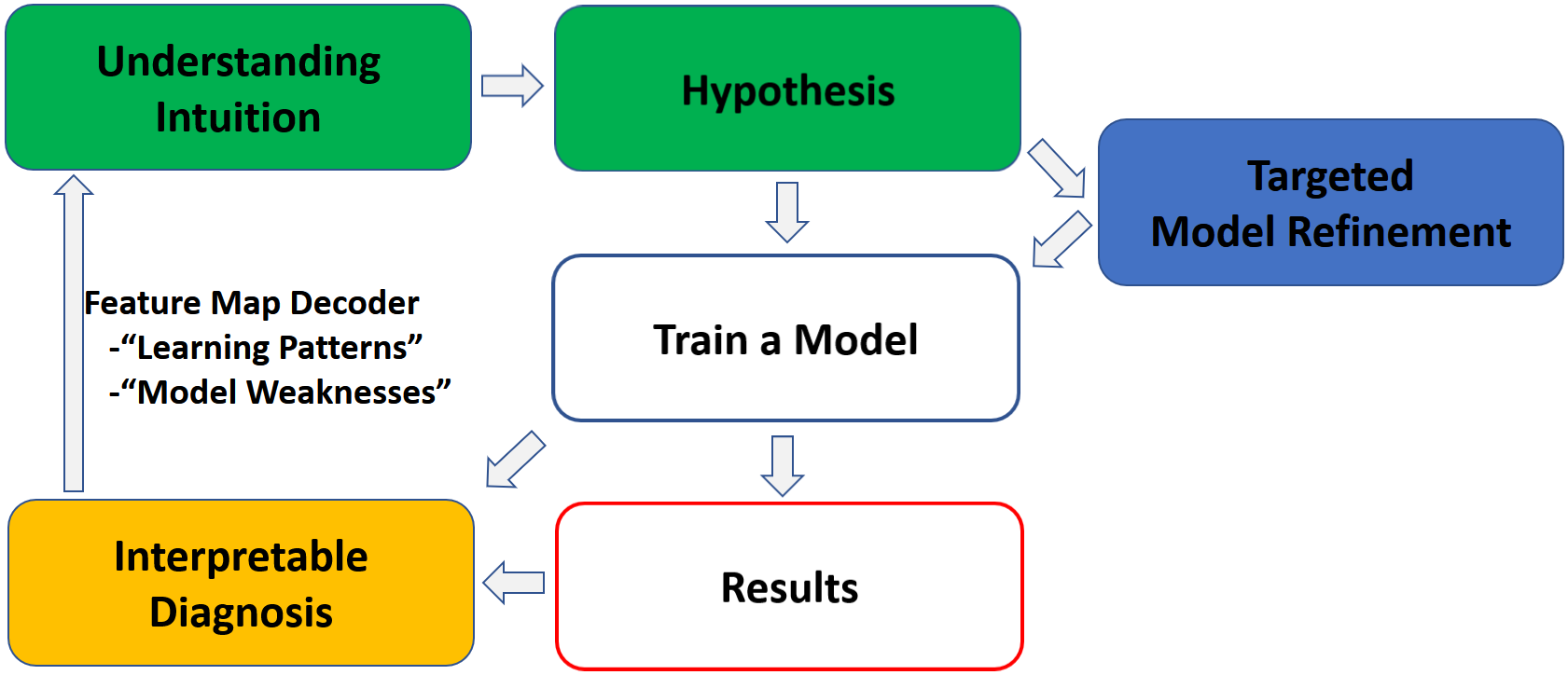}
\end{center}
 \caption{The 'Train, Diagnose and Fix' paradigm provides means of systematic model interpretation which enables targeted model refinement strategies for learning stronger recognition models. All figures are best viewed in color.}
\label{fig:thumb_nail}
\end{figure}

Despite the remarkable recognition power of recent deep learning methods, our ability to clearly explain the inner workings of deep networks has been unsatisfying \cite{Authors14,Authors15,Authors25,Authors26}. Fundamentally, without the potential to fully understand and interpret deep learning models, the development cycle for model training mostly reduces down to 'trial-and-error' \cite{Authors13}. In this light, we propose a more systematic and closed-loop approach to learning models through model diagnosis and targeted refinement. We wish to take a step towards a 'train-diagnose-and-fix' paradigm and move away from an exhaustive 'train-and-error' routine. As Figure \ref{fig:thumb_nail} suggests, interpretable model diagnosis enables us to form better understanding of the problem of interest, reach a deeper understanding of how the model is learning, identify model's weaknesses and ultimately form better hypotheses to train stronger models. In this work, we propose two key components of the 'train-diagnose-fix' framework: the \textit{diagnosis} tool and an approach to \textit{fix} a model via targeted model refinement.

In human action recognition from skeleton data in particular, despite the significant discriminative power of LSTM-based Recurrent Neural Networks architectures \cite{Authors1,Authors2,Authors3,Authors4,Authors5,Authors8,Authors10}, the growing complexity of recent models hinders our ability to intuitively interpret them. A recently proposed Res-TCN \cite{Authors7} is specifically designed to improve the interpretability of model parameters while efficiently learning a discriminative spatio-temporal representation for human action recognition with a convolutional neural network (CNN) based approach.

However, the work of \cite{Authors7} limits its narrative on the interpretation of the model parameters and expand only a cursory exploration to hidden representations. Moreover, the adopted visualization technique is relatively primary, which lends no sufficient support to the proposed interpretation and leaves it mostly intuitive. With the presented visualization results, little extra information can be further extracted for diagnosis and targeted refinement. In this light, we propose a more systematic approach for model interpretation and the means for model refinement given the findings from the diagnosis. The main contribution of this work is three fold:
\begin{itemize}[leftmargin=*,nosep]
\item 
A model diagnosis tool, the Feature Map Decoder, is proposed for systematic interpretation of model parameters and hidden representations. Feature Map Decoder visualizes the hidden representations in the same domain as the input. The proposed Feature Map Decoder effectively exposes the learning patterns and weaknesses of a trained model. In our work, the diagnosis uncovered that the model learned to implicitly transform the input sequence into a more discriminative view point over the layers of the network and the characteristic spatio-temporal portions of the input is gradually accentuated throughout the layers.
\item 
A targeted refinement technique is proposed to encode cognitive-driven prior knowledge into a data-driven deep learning model. A Targeted Attention (TA) stream is introduced as part of the refinement technique to force the model to learn additional representation over the specific dimensions of the input. Such set of dimensions where the model struggles the most is identified by the interpretable diagnosis. A \textit{Pipe} architecture is introduced to effectively fuse the information learned in the TA stream to the original model. 
\item
Following the 'train-diagnose-and-fix' pipeline, the resulting Multi-Stream Residual Temporal Convolutional Network (MS-Res-TCN) achieves the state-of-the-art performance on NTU-RGBD dataset \cite{Authors2}. 
\end{itemize}

\section{Related Work}

In this section, we first provide a brief review on relevant 3D skeleton based human action recognition approaches. We then extend our survey to current literature on interpretation of modern deep neural networks.

\subsection{3D Skeleton Based Human Action Recognition}

Given the time-series nature of the data, models that employ LSTM-based Recurrent Neural Networks have become an active architecture of the research. HBRNN \cite{Authors1} divides the whole skeleton into five parts. The parts are first fed separately into five bidirectional recurrent neural subnets, and then fused hierarchically as the layers go deeper. In Part-Aware LSTM model \cite{Authors2}, part-based cells are introduced to keep the context of each of the five body parts independent. The final classifier is learned over these independent sub predictions. In the work of \cite{Authors3}, rather than pre-defining the joint groups as in \cite{Authors1,Authors2}, a co-occurrence exploration process is applied to achieve flexible automatic co-occurrence mining. The work of \cite{Authors4} extends the analysis to the spatial domain to substitute the concatenation of the skeleton joint information as in previous methods \cite{Authors1,Authors2,Authors3}. In this fashion, a spatio-temporal LSTM model with trust gates is introduced to concurrently learn both the spatial structure of the joints and the temporal dependencies among frames. Similarly, in STA-LSTM \cite{Authors5}, a spatial attention module is designed to automatically select dominant joints within each frame, and a temporal attention module is designed to assign different degrees of importance among the frames.

In action segmentation and detection, a new class of temporal models, Temporal Convolutional Network (TCN) \cite{Authors6}, is proposed to capture long-range temporal dependencies, which outperforms other LSTM-based Recurrent Networks in both accuracy and training time. The work of \cite{Authors7} employs a re-designed TCN model (Res-TCN) with residual connections  \cite{Authors8,Authors9} to skeleton based human action recognition task. The proposed Res-TCN also learns both spatial and temporal attention for human action recognition. The authors of \cite{Authors7} argue that Res-TCN is specifically designed to enhance the interpretability of model parameters and features compared to other recurrent models. However, the work of \cite{Authors7} remains mostly intuitive and unsystematic as opposed to our work.

\subsection{On Neural Network Interpretability}

Recently, there has been an increase in efforts to interpret deep neural networks \cite{Authors11,Authors12,Authors13,Authors14,Authors15,Authors16,Authors17,Authors18,Authors19,Authors20,Authors21,Authors22}. The behavior of a complex black-box deep neural network can be understood by generating a locally faithful interpretable model around the test point and observing the output \cite{Authors15}. The authors of \cite{Authors23,Authors24} perturb the test point and observe how the prediction of the model changes. Additionally, \cite{Authors13,Authors16,Authors17,Authors19} attempt to visualize the synthesized inputs to the network that maximize the activations of hidden units.

Another important class of approaches attempts to understand a CNN-based deep learning model by looking directly into the hidden representations with well-designed visualization techniques, which is most relevant to our current work. In the work of \cite{Authors11}, the proposed Network Dissection method performs a binary segmentation on every feature map with respect to every visual concept in Broden Dataset \cite{Authors11}. The intersection-over-union (IoU) score for every segmentation task is computed. A visual concept is matched to every neuron in the network by ranking the IoU scores. The work of \cite{Authors12} proposes an optimization method to compute an approximate inverse of the hidden representations. Note that the motivation of the work of \cite{Authors12} is to reconstruct the input image from the hidden representations as accurate as possible. We argue that the proposed optimization method has no interpretable basis as opposed to our Feature Map Decoder which can be viewed as an inverse computation of a single forward pass in the neural network.

\section{Overview of Residual Temporal Convolutional Networks}

\begin{figure}[h]
\begin{center}

\includegraphics[height=7.5cm]{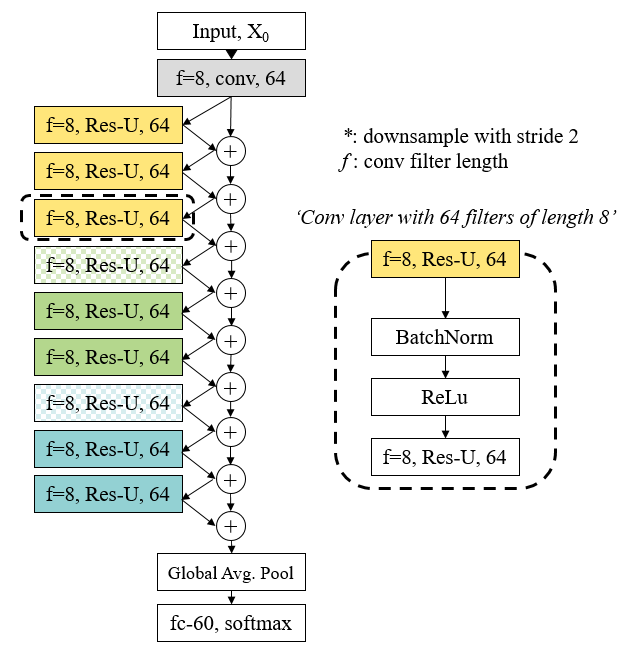}
\end{center}
   \caption{Res-TCN model architecture. Except the first convolution layer (in gray), the model consists of stacked residual units.}
\label{fig:Res-TCN}
\end{figure}

In this section, we provide a brief overview of the Residual Temporal Convolutional Network \cite{Authors7} (Res-TCN) architecture (Figure \ref{fig:Res-TCN}). Res-TCN is an extension of TCN \cite{Authors6} with residual connections with identity mappings as proposed in \cite{Authors8,Authors9}. Res-TCN can naturally model a wide range of time-series data. Let $X_0 \in \mathbb R^{T \times D}$ be the input that represents a time-series data of temporal length $T$ with $D$ dimensional feature per time step. Res-TCN hierarchically stacks building blocks called \textit{Residual Units}. Each unit in layer $l \geq 2$ performs the following computation:
\begin{equation} \label{eq:1}
 X_{l} = X_{l-1} + F(W_l,X_{l-1})
\end{equation}
\begin{equation} \label{eq:2}
F(W_l,X_{l-1}) = W_l*\sigma(X_{l-1})
\end{equation} 
$F$ denotes the residual unit. For the $l$-th layer, $X_{l-1}$ denotes the input, $W_l$ is the set of learnable parameters and $\sigma$ is batch normalization \cite{Authors27} followed by a ReLU activation function. Note that the first convolution layer in Res-TCN operates on raw time-series input and the resulting activation map, $X_1$, is passed on to subsequent layers without going through a normalization or an activation layer. Given a Res-TCN with $N$ residual units, the hidden representation after $N$ residual units is:
\begin{equation} \label{eq:3}
X_N = X_1 + \sum_{i=2}^{N} W_i*\sigma(X_{i-1})
\end{equation}
\begin{equation} \label{eq:4}
X_1 = W_1X_0
\end{equation}
The set of filters in $W_1$ and the resulting activation map, $X_1$, are readily interpretable given that each dimension of $X_0$ has a semantic meaning associated with it. For example, a sequence of robot kinematics or 3D skeletons are such input domains. This property of Res-TCN is the driving motivation of the formulation of the proposed Feature Map Decoder in Section \ref{section:fmd}.

For prediction, we apply global average pooling after the last merge layer across the entire temporal sequence and attach a softmax layer with number of neurons equal to number of classes.   

\section{Interpretation of Residual Temporal Convolutional Networks}
\begin{figure*}[ht]
\begin{center}
	\includegraphics[width=1.00\textwidth,height=4cm]{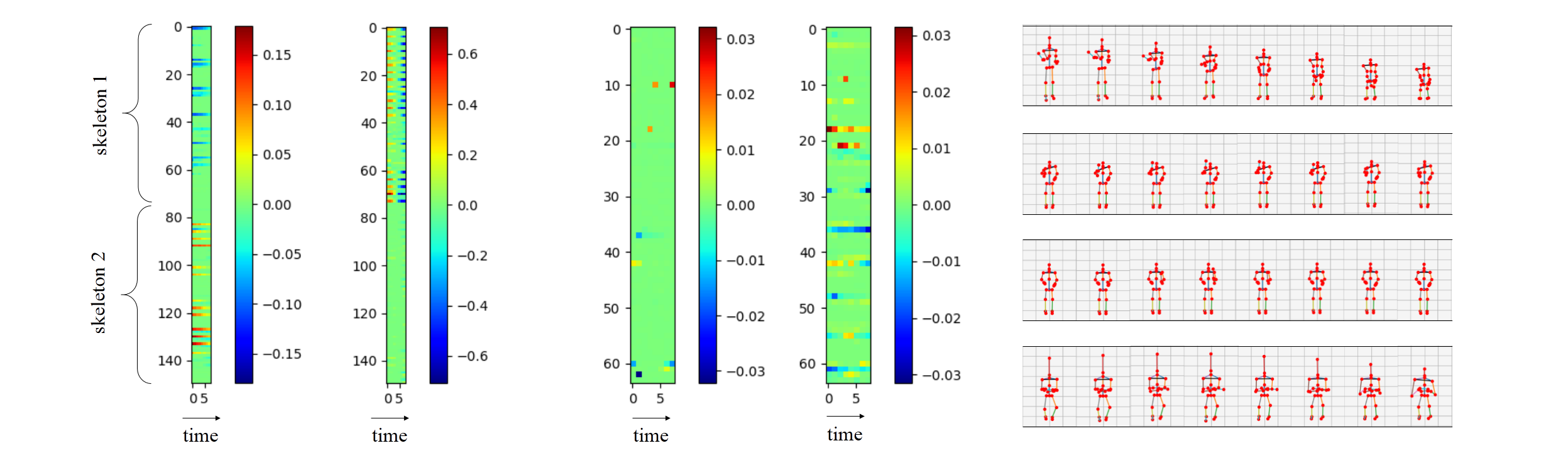}
\end{center}
   \caption{\textbf{Left:} The first two filters visualize the temporal filters in the first layer. \textbf{Middle:} The two filters in the middle are examples of deeper-layer filters. \textbf{Right:} First layer conv filters plotted directly as moving skeletons. This shows that such filters can be thought of as \textit{motion primitives}.}
\label{fig:mp}
\end{figure*}
In this section, we first propose an interpretation of model parameters and hidden representations on an intuitive level. The insight gained from such analysis leads to the formulation of our Feature Map Decoder. The decoded skeleton sequences from the Feature Map Decoder expose the characteristic learning patterns of Res-TCN. Finally, we introduce a targeted refinement technique, which effectively encodes the newly obtained knowledge gained by interpretation analysis into a deep learning model.

If all dimensions of the input have interpretable semantic meaning associated with them, then we can directly understand every parameter of the first convolution layer \cite{Authors7}. For example, given a sequence of 3D human joints as input, the first-layer filters can be directly interpreted as filters that learn to detect snippets of joint motion. Each dimension of a first-layer filter, $W^{(i)}_1 \in \mathbb R^{f_1 \times D}$, has a corresponding semantic joint associated with it, where $f_1$ denotes the temporal filter length and $D$ is the dimension of the input per frame. We will refer the snippets of motion learned in the first layer as \textit{motion primitives} in the following narrative. 

Filters belonging to deeper layers can also be interpreted. Given the model architecture of Res-TCN, the convolution filters in deeper layers act as gating operations. For example, the hidden representation after the first residual unit is:
\begin{equation} \label{eq:5}
X_2 = X_1 + W_2*\sigma(0,X_1)
\end{equation}
where $X_1$ is the output of the first convolution layer, $W_2$ represents the set of filters in the second convolution layer. $X_1$ represents to what extent and how the motion primitives in $W_1$ are distributed in the input $X_0$. Following this logic and observing Equation \ref{eq:5}, we view $W_2$ as a gate that learns to find a discriminative weighted combination of motion primitives encoded in $X_1$. Given Equation \ref{eq:3}, the same argument can be made for all subsequent layers. Figure \ref{fig:mp} visualizes the conv filters in the first and deeper layers. The filters of the first layer show a continuous and smooth changing motion-like structure for each dimension (corresponding to a semantic joint) across time, while filters of deeper layers show a discrete and sparse gate-like structure. Note that the channel dimensionality of conv1 filters are equivalent to the dimensionality of the input skeleton. Then, to be more explicit, we provide visualization of the first-layer filters as a sequence of moving skeletons (Figure \ref{fig:mp}), which in fact validates that these filters can be thought of as motion primitives. We simply frame-wise map each dimension of a filter back to its corresponding semantic joint, and then add back the mean skeleton which is subtracted in the training process.

Let us now extend our analysis to hidden representations. A convolution operation with a filter produces a high positive response over the regions of the input where the shape of the filter is highly correlated with the input region. If the input is a sequence of 3D skeletons, a filter in $W_1$ (motion primitive) produces a high positive convolution response in temporal locations of $X_0$ where $X_0$ shows a similar motion pattern defined in that particular filter. Let the activation map of layer $l$ be $X_l \in \mathbb R^{T \times N_l}$, where $N_l$ denotes the number of filters in layer $l$ and $T$ represents the input length. $x^{(i)}_l(t)$ can be interpreted as the $i$-th motion primitive's activation response for the input sequence $X_0$ at frame $t$.

\subsection{Feature Map Decoder}
\label{section:fmd}

\begin{figure*}[ht]
\begin{center}
\includegraphics[width=1.05\textwidth,height=5cm]{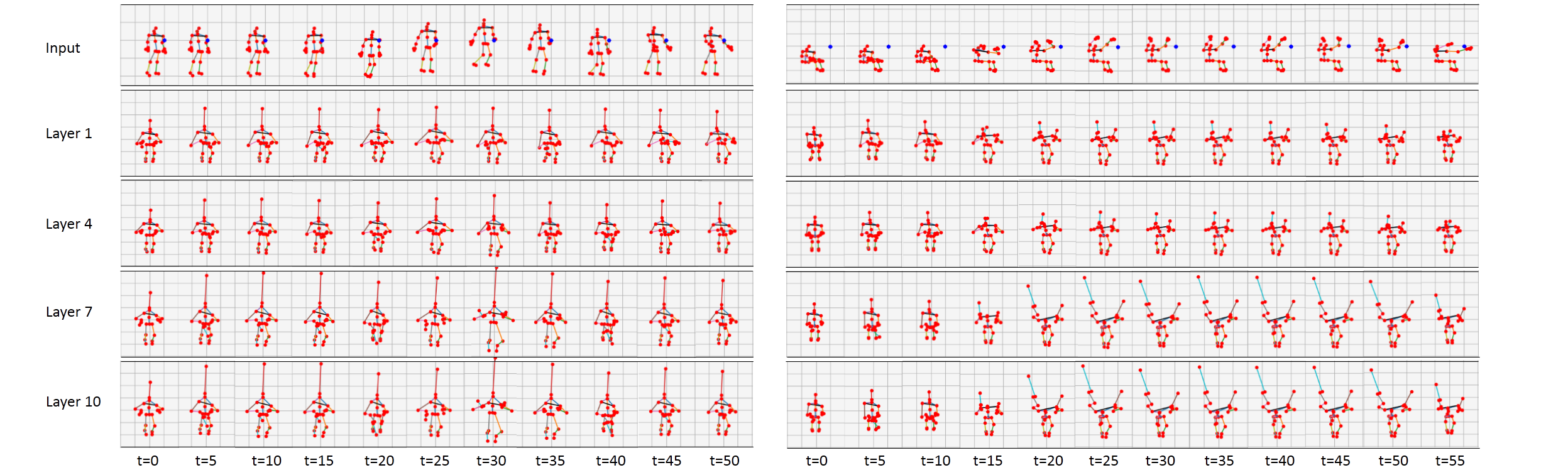}
\end{center}
\caption{Visualization of decoded skeleton sequences from layers 1,4,7,10. The visualization shows that discriminative motion patterns are gradually emphasized through the layers. \textbf{Left:} An example sequence of action class \textit{jump up}. \textbf{Right:} An example sequence of action class  \textit{cheer up}.}
\label{fig:magnify}

\begin{center}
\includegraphics[width=1.05\textwidth,height=5cm]{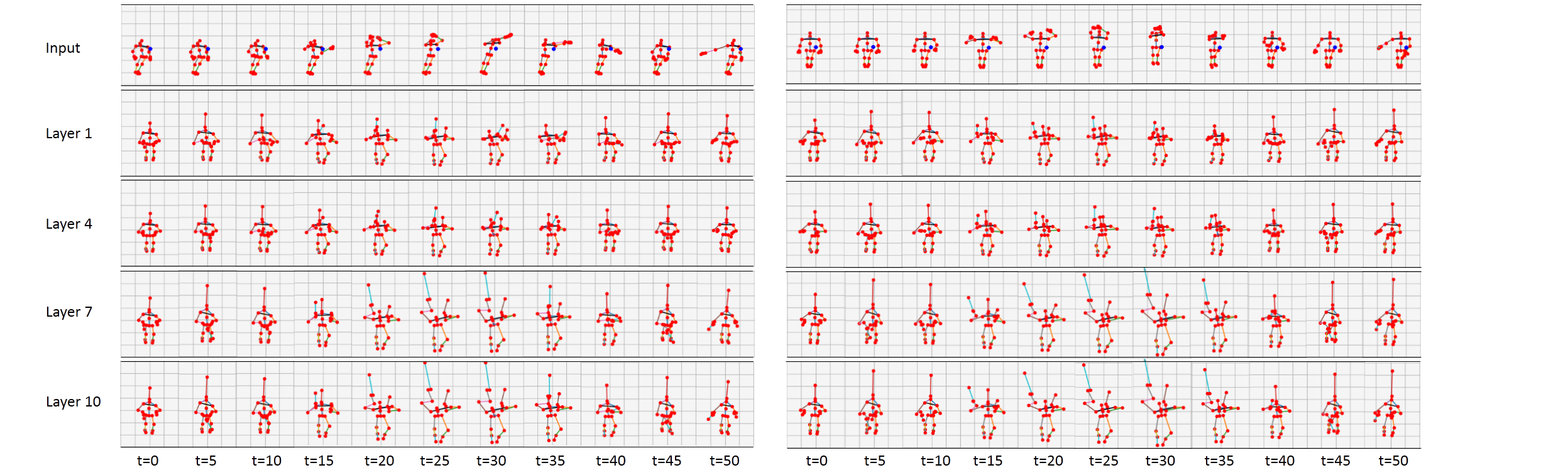}
\end{center}
\caption{Analysis of the model through Feature Map Decoder reveals that the model has achieved view-point invariance. \textbf{Left:} An example sequence of action class \textit{throw} viewed from the side. \textbf{Right:} An example sequence of action class \textit{throw} viewed from the front.}
\label{fig:view}
\end{figure*}
As mentioned above, the first-layer filters learn a set of motion primitives; the deeper-layer filters can be interpreted as gates, which hierarchically fuse the motion primitives learned in the first layer; the activation map of each layer represents a temporal distribution of the learned motion primitives' responses. Therefore, we argue that using the set of motion primitives, the hidden representations of the model can be converted back to a sequence of 3D skeletons performing some combination of these motion primitives. This leads to our proposed Feature Map Decoder formulation.

Let $\hat{X_l} \in \mathbb R^{T \times D}$ be the skeleton sequence decoded from the activation map of the $l$-th layer, $X_l \in \mathbb R^{T \times N_l}$. We first consider a simplified case when the number of filters is constant across all layers, i.e. $N_l = N_c$, $\forall l \geq 1$. For each filter of the first layer, $W^{(i)}_1 = \{w^{(i)}_1(t)\}^{f_1}_{t=1}$, where $f_1$ denotes the filter length of the first-layer filters, the compressed version of the filter is formulated as:
\begin{equation} \label{eq:6}
\hat{W}^{(i)}_1 = {(w^{(i)}_1(f_1/2) + w^{(i)}_1(f_1/2+1))}/2
\end{equation}
where $\hat{W}^{(i)}_1 \in \mathbb R^{1 \times D}$. We are computing the mean of the inner most two time steps of the filter following the observation that the most characteristic motion patterns appear near the middle time steps of the temporal filters. In this fashion, motion primitives of length $f_1$ learned by the first-layer filters are compressed into a single time step. Then, the filter responses in each time step are used to compute the weighted sum of    $\{\hat{W}^{(i)}_1\}^{N_c}_{i=1}$:
\begin{equation} \label{eq:7}
\hat{x}_l(t) = \sum_{i=1}^{N_c}\hat{W}^{(i)}_1 \times x^{(i)}_l(t)
\end{equation}
where $x^{(i)}_l(t)$ is a $i$-th channel response at time step $t$ of $X_l$. $\hat{x}_l(t) \in \mathbb R^{1 \times D}$ is the decoded skeleton at time step $t$ computed as a weighted combination of all compressed motion primitives. We repeat this process for all $t \in (0,T)$ and temporal concatenation of all $\{\hat{x}_l(t)\}^{T}_{t=1}$ yields a full decoded sequence $\hat{X_l} \in \mathbb R^{T \times D}$. Note that in the training stage, a mean skeleton has been subtracted from input samples. Hence, we add the mean skeleton back to the decoded skeleton sequence to maintain mathematical consistency.

Now, let us extend our discussion to the circumstance when the number of filters changes between layers. In the Res-TCN architecture, layers in the same block share equal number of filters. Let $N_{1,2,3,4} = N_{c1}$, $N_{5,6,7} = N_{c2}$, $N_{8,9,10} = N_{c3}$. Note that when the number of filters changes between the residual units ($l=5,8$), the identity mapping connection requires an additional convolution for channel shape matching purposes:
\begin{equation} \label{eq:8}
X_l = \tilde{W}_l X_{l-1} +  F(W_l,X_{l-1}), \quad l=5,8
\end{equation}
\begin{equation} \label{eq:9}
F(W_l,X_{l-1}) = W_l*\sigma(X_{l-1})
\end{equation}
where $\tilde{W}_l$ denotes the convolution layer attached to the identity mapping connection.

\begin{figure*}[ht]
\begin{center}
\includegraphics[width=1.05\textwidth,height=5cm]{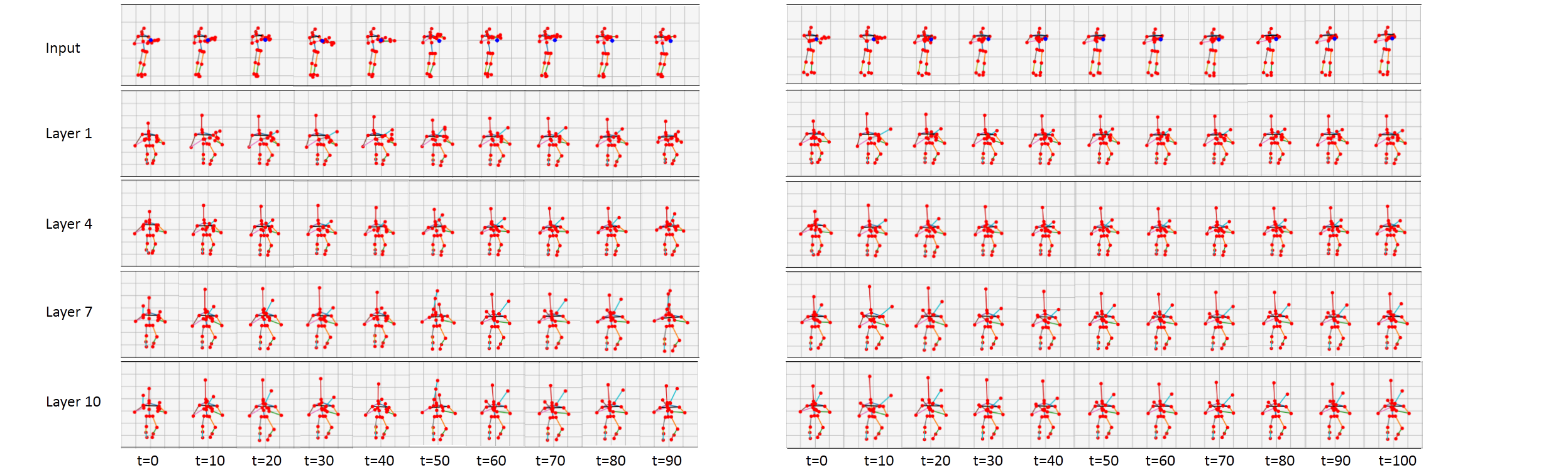}
\end{center}
\caption{Examples of decoded skeleton sequences output from the Feature Map Decoder, which exposes Res-TCN's failure to capture and model detailed hand joints motion. \textbf{Left:} A case study of action class \textit{reading}. \textbf{Right:} A case study of action class \textit{writing}.}
\label{fig:hand}
\end{figure*}
For the first layer and the units in Block-A, no additional decoding step is needed because $N_1 = N_{c1}$ and the feature map decoding procedure follows the steps described in Equations \ref{eq:6} and \ref{eq:7}. For the layers in Block-B, one-step retrieval needs to be performed to reshape the decoded sequence from $X_l \in \mathbb R^{T \times N_{c2}}$ to $X^{'}_l \in \mathbb R^{T \times N_{c1}}$. The retrieval procedure is equivalent to steps listed in Equation \ref{eq:6} and \ref{eq:7} but the activations are first decoded using a set of $\hat{W}^{(i)}_5 \in \mathbb R^{1 \times N_{c1}}$ compressed filters instead of a set of $\hat{W}^{(i)}_1 \in \mathbb R^{1 \times D}$ compressed filters. The retrieval is computed as follows:
\begin{multline} \label{eq:10}
\hat{x}^{'}_l(t) = \sum_{i=1}^{N_{c_2}} [ \hat{W}^{(i)}_5 \times (x^{(i)}_l(t) - x^{(i)}_5(t)) \\
+ \tilde{W}^{(i)}_5 \times x^{(i)}_5(t)], \quad l=5,6,7
\end{multline}
$\hat{W}^{(i)}_5 \in \mathbb R^{1 \times N_{c1}}$ is the $i$-th compressed filter of the residual unit of layer $l=5$. $\tilde{W}^{(i)}_5 \in \mathbb R^{1 \times N_{c1}}$ is the $i$-th conv filter of temporal length $1$ in layer $l=5$. $x^{(i)}_l(t) \in \mathbb R$ and $x^{(i)}_5(t) \in \mathbb R$ are the $i$-th filter's activation response in layer $l$ and the fifth layer respectively, at time step $t$. $\hat{x}^{'}_l(t) \in \mathbb R^{1 \times N_{c1}}$ is the retrieved activation map at time step $t$ and the temporal concatenation across all time steps yields the intermediate result $X^{'}_l \in \mathbb R^{T \times N_{c1}}$. Then, the steps listed in Equation \ref{eq:6} and \ref{eq:7} are repeated with $X^{'}_l$ to produce the final decoded sequence of $X_l \in \mathbb R^{T \times N_D}$.

Similar operations are performed for hidden representations in Block-C. For such activations, a set of $\hat{W}^{(i)}_8 \in \mathbb R^{1 \times N_{c2}}$ compressed filters first retrieve the sequence to activation shapes of Block-B. Then, the decoding procedure follows the same steps as the activations in Block-B.

The Res-TCN down samples the temporal dimension by a factor of $2$ after each block. To simplify the narrative, we have omitted the interpolation step required to up-sample the temporal dimension of the retrieved decoded sequences. 

\subsection{Targeted Model Refinement}
\label{refinement}
\begin{figure*}
\begin{center}
\includegraphics[width=0.90\textwidth,height=6cm]{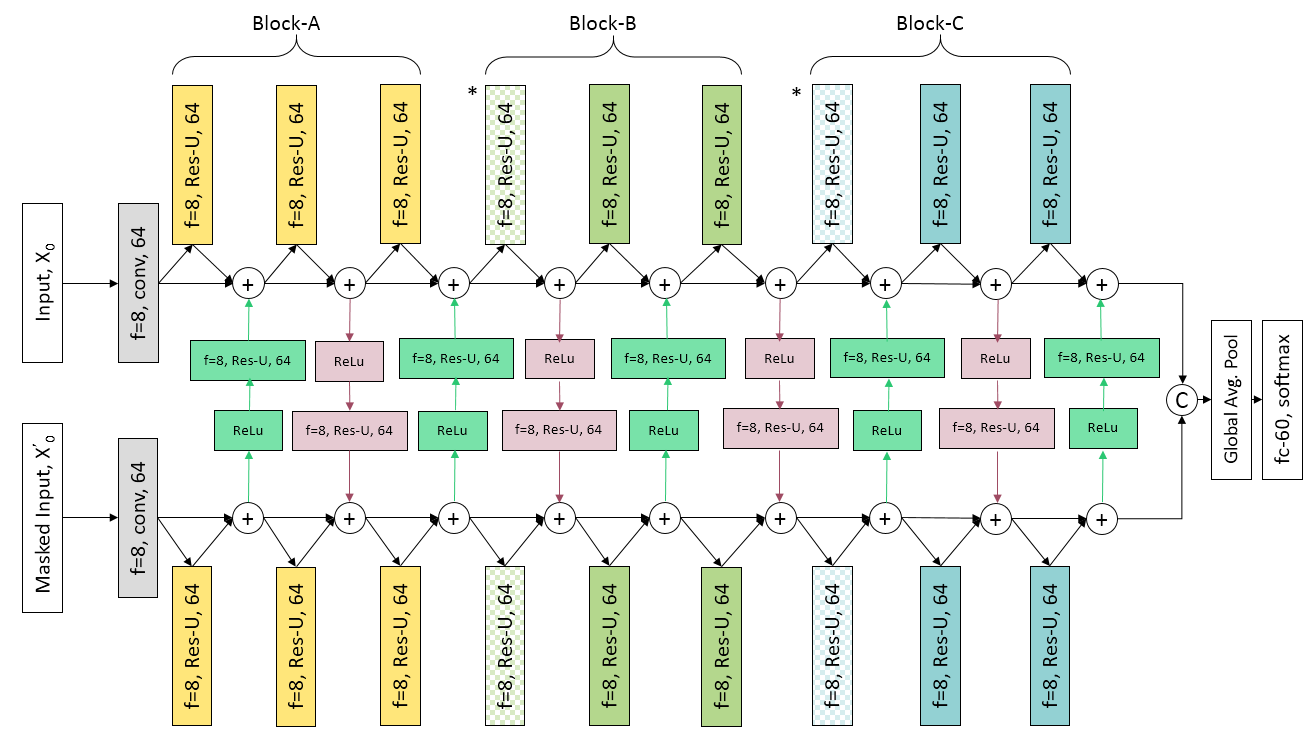}
\end{center}
   \caption{Targeted Model Refinement technique yields MS-Res-TCN architecture. The TA stream receives a masked input sequence and the main stream receives the original input.}
\label{fig:MS-Res-TCN}
\end{figure*}
Through intuitive visualization, the proposed Feature Map Decoder in Section \ref{section:fmd} exposes the learning patterns and weaknesses of the model. Our diagnosis findings will be discussed in more depth in Section \ref{section:exp_interp} but as a running example, we identified that our model struggles to detect very fine-grained motion patterns of the hands. Given such finding, we introduce Targeted Attention (TA) stream as an approach to explicitly address the weaknesses identified in the diagnosis step. The TA stream receives input, $X^{'}_0 \in \mathbb R^{T \times D}$, from a masked version of the original input where all dimensions are masked except the targeted dimensions. For the example discussed above, $X^{'}$ contains only the time-series information from the input dimensions corresponding to the hands and all other dimensions are masked. The explicitly targeted representation of the TA stream is fused with the original stream at every merge layer (the \textit{Pipe}) so that the representation learning of the two streams is mutually interactive. A convolution layer followed by a ReLU activation is employed on each \textit{Pipe} to control how much information can be transferred between two streams. Figure \ref{fig:MS-Res-TCN} depicts the resulting architecture, Multi-Stream Res-TCN (MS-Res-TCN).

Each unit in layer $l \geq 2$ performs the following computation:
\begin{equation}
X_l = X_{l-1} + F(W_l,X_{l-1}) + H(W^P_{l},X^{'}_l)
\end{equation}
\begin{equation}
X^{'}_l = X^{'}_{l-1} + F(W^{'}_l,X^{'}_{l-1}) + H(W^P_{l},X_l)
\end{equation}
\begin{equation}
F(W_l,X_{l-1}) = W_l*\sigma(X_{l-1})
\end{equation}
\begin{equation}
F(W^{'}_l,X^{'}_{l-1}) = W^{'}_l*\sigma(X^{'}_l{-1})
\end{equation}
\begin{equation}
H(W^P_{l},X^{'}_l) = \left\{
\begin{array}{lr}
\sigma(W^P_{l}*X^{'}_l) & {l=2k}\\
0 & {l=2k+1}
\end{array} \right.
\end{equation}
\begin{equation}
H(W^P_{l},X_l) = \left\{
\begin{array}{lr}
0 & {l=2k}\\
\sigma(W^P_{l}*X_l) & {l=2k+1}
\end{array} \right.
\end{equation}
where $H$ denotes the \textit{Pipe} operation, $W^{'}_l$ is a set of convolution filters in the $l$-th layer of the TA stream, $W^P_l$ represents the learnable convolution filter parameters in the $l$-th $Pipe$.

The outputs from the two streams are concatenated. The concatenated data is then fed into the same prediction layer (global average pool followed by a softmax layer).

\section{Experiments}
In this section, we analyze our findings from the model diagnosis and validate the effectiveness of our targeted refinement technique on 3D skeleton based human activity recognition benchmark of NTU RGB+D \cite{Authors2}. 

\subsection{Dataset and Settings}

\textbf{NTU RGB+D Dataset (NTU)}:
The NTU RGB+D dataset \cite{Authors2} is currently the largest 3D skeleton human action recognition dataset, which contains 56880 video samples of 60 action classes collected from 40 distinct subjects. For each setup, a subject's action is recorded from three cameras of same height but from different viewing angles: $-45^{\circ}$, $0^{\circ}$ and $+45^{\circ}$. The dataset provides two criteria of validation: Cross-Subject (CS) and Cross-View (CV) evaluations. The provided skeletons have 25 joints.

\textbf{Implementation Details}:
We do not perform view-point or scale normalization \cite{Authors1,Authors2,Authors3} and use the raw $(X,Y,Z)$ coordinates. Per time-step feature is 120 dimensional ($25$ joints, $2$ at most actors, $3$ xyz positions per joint). We perform all our experiments on a Nvidia K80 GPU with Keras2.0 \cite{Authors29} using a TensorFlow \cite{Authors31} backend. The learning rate is initially set to $0.01$ and then decreases by a factor of $10$ when the testing loss plateaus. We train with Stochastic Gradient Descent with all standard parameters and with a batch size of 128. L-1 regularizer with a weight of $1e-4$ is applied to all convolution layers and Dropout \cite{Authors30} ($p=0.5$) is applied after every ReLU operation. The implementation and converged model weights will be made publicly available. (The link will not shared for this submission for the sake of double-blind review).

\begin{table}[b]
\centering
\caption{Comparison to other state-of-the-art models on NTU.}
\label{table:NTU}
\begin{tabular}{|c|c|c|}
\hline
\rowcolor[HTML]{C0C0C0} 
\textbf{Model}                     & \textbf{Cross-Subjects} & \textbf{Cross-Views} \\ \hline
Dynamic Skeletons \cite{Authors33}                 & 60.2                    & 65.2                 \\ \hline
HBRNN \cite{Authors1}                             & 59.1                    & 64.0                 \\ \hline
Deep LSTM \cite{Authors2}                         & 60.7                    & 67.3                 \\ \hline
P-LSTM \cite{Authors2}                            & 62.9                    & 70.3                 \\ \hline
Trust Gate \cite{Authors4}                        & 69.2                    & 77.7                 \\ \hline
STA-LSTM \cite{Authors5}                          & 73.4                    & 81.2                 \\ \hline
JTM \cite{Authors35}                               & 76.3                    & 81.1                 \\ \hline
Pose Conditioned STA \cite{Authors34}               & 77.1                    & 84.5                 \\ \hline
Res-TCN baseline \cite{Authors7}                  & 74.3                    & 83.1                 \\ \hline
\textit{\textbf{MS-ResTCN, our model}} & \textbf{79.0}           & \textbf{86.6}        \\ \hline
\end{tabular}
\end{table}

\begin{table*}[ht]
\caption{Detailed performance comparison between the Res-TCN baseline and MS-Res-TCN with Cross-Views evaluation. Highlighted in bold are action classes that show meaningful increase in performance.}
\label{tabel:NTU-detail}
\resizebox{\textwidth}{!}{%
\begin{tabular}{lllllllllll}
\hline
           & \cellcolor[HTML]{EFEFEF}\textit{\textbf{\begin{tabular}[c]{@{}l@{}}Drink\\ Water\end{tabular}}}                           & \cellcolor[HTML]{EFEFEF}\begin{tabular}[c]{@{}l@{}}Eat Meal\\ / Snack\end{tabular}     & \cellcolor[HTML]{EFEFEF}\begin{tabular}[c]{@{}l@{}}Brushing\\ Teeth\end{tabular}                & \cellcolor[HTML]{EFEFEF}\textit{\textbf{\begin{tabular}[c]{@{}l@{}}Brushing\\ Hair\end{tabular}}} & \cellcolor[HTML]{EFEFEF}Drop                                                                                    & \cellcolor[HTML]{EFEFEF}Pick Up                                                                     & \cellcolor[HTML]{EFEFEF}Throw                                                                                     & \cellcolor[HTML]{EFEFEF}\begin{tabular}[c]{@{}l@{}}Sitting\\ Down\end{tabular}          & \cellcolor[HTML]{EFEFEF}\begin{tabular}[c]{@{}l@{}}Standing\\ Up\end{tabular}                            & \cellcolor[HTML]{EFEFEF}\textit{\textbf{Clapping}}                                                                              \\ \hline
\rowcolor[HTML]{FFFFFF} 
Baseline   & \textbf{0.7839}                                                                                                  & 0.7225                                                                                 & 0.8168                                                                                          & \textbf{0.7915}                                                                                   & 0.8892                                                                                                          & 0.9098                                                                                              & 0.9263                                                                                                            & 0.9587                                                                                  & 0.9832                                                                                                   & \textbf{0.6614}                                                                                                        \\ \hline
\rowcolor[HTML]{FFFFFF} 
MS-Res-TCN & \textbf{0.8418}                                                                                                  & 0.7690                                                                                 & 0.8354                                                                                          & \textbf{0.8861}                                                                                   & 0.9177                                                                                                          & 0.9335                                                                                              & 0.9589                                                                                                            & 0.9714                                                                                  & 0.9842                                                                                                   & \textbf{0.7373}                                                                                                        \\ \hline
           & \cellcolor[HTML]{EFEFEF}\textit{\textbf{Reading}}                                                       & \cellcolor[HTML]{EFEFEF}Writing                                                        & \cellcolor[HTML]{EFEFEF}\begin{tabular}[c]{@{}l@{}}Tear Up\\ Paper\end{tabular}                 & \cellcolor[HTML]{EFEFEF}\begin{tabular}[c]{@{}l@{}}Wear\\ Jacket\end{tabular}                     & \cellcolor[HTML]{EFEFEF}\begin{tabular}[c]{@{}l@{}}Take Off\\ Jacket\end{tabular}                               & \cellcolor[HTML]{EFEFEF}\textit{\textbf{\begin{tabular}[c]{@{}l@{}}Wear \\ Shoe\end{tabular}}}      & \cellcolor[HTML]{EFEFEF}\textit{\textbf{\begin{tabular}[c]{@{}l@{}}Take Off \\ Shoe\end{tabular}}}                & \cellcolor[HTML]{EFEFEF}\begin{tabular}[c]{@{}l@{}}Wear On\\ Glasses\end{tabular}       & \cellcolor[HTML]{EFEFEF}\begin{tabular}[c]{@{}l@{}}Take off\\ Glasses\end{tabular}                       & \cellcolor[HTML]{EFEFEF}\textit{\textbf{\begin{tabular}[c]{@{}l@{}}Put On a \\ Hat/Cap\end{tabular}}}         \\ \hline
Baseline   & \textbf{0.3759}                                                                                         & 0.4610                                                                                 & 0.8174                                                                                          & 0.9356                                                                                            & 0.9244                                                                                                          & \textbf{0.7280}                                                                                     & \textbf{0.6416}                                                                                                   & 0.8256                                                                                  & 0.9196                                                                                                   & \textbf{0.8143}                                                                                               \\ \hline
MS-Res-TCN & \textbf{0.5429}                                                                                         & 0.4603                                                                                 & 0.8987                                                                                          & 0.9778                                                                                            & 0.9492                                                                                                          & \textbf{0.8280}                                                                                     & \textbf{0.7524}                                                                                                   & 0.8544                                                                                  & 0.9082                                                                                                   & \textbf{0.9270}                                                                                               \\ \hline
           & \cellcolor[HTML]{EFEFEF}\textit{\textbf{\begin{tabular}[c]{@{}l@{}}Take Off a \\ Hat/Cap\end{tabular}}} & \cellcolor[HTML]{EFEFEF}Cheer Up                                                       & \cellcolor[HTML]{EFEFEF}\textit{\textbf{\begin{tabular}[c]{@{}l@{}}Hand\\ Waving\end{tabular}}} & \cellcolor[HTML]{EFEFEF}\begin{tabular}[c]{@{}l@{}}Kicking\\ Something\end{tabular}               & \cellcolor[HTML]{EFEFEF}\textit{\textbf{\begin{tabular}[c]{@{}l@{}}Put Something\\ Inside Pocket\end{tabular}}} & \cellcolor[HTML]{EFEFEF}Hopping                                                                     & \cellcolor[HTML]{EFEFEF}Jump Up                                                                                   & \cellcolor[HTML]{EFEFEF}\textit{\textbf{\begin{tabular}[c]{@{}l@{}}Make a \\ Phone Call\end{tabular}}}    & \cellcolor[HTML]{EFEFEF}\begin{tabular}[c]{@{}l@{}}Playing with\\ Phone\end{tabular}                     & \cellcolor[HTML]{EFEFEF}\textit{\textbf{\begin{tabular}[c]{@{}l@{}}Typing on\\ Keyboard\end{tabular}}}        \\ \hline
Baseline   & \textbf{0.8810}                                                                                         & 0.9589                                                                                 & \textbf{0.7652}                                                                                 & 0.8877                                                                                            & \textbf{0.7535}                                                                                                 & 0.9715                                                                                              & 0.9835                                                                                                            & \textbf{0.7877}                                                                                  & 0.6665                                                                                                   & \textbf{0.6335}                                                                                               \\ \hline
MS-Res-TCN & \textbf{0.9494}                                                                                         & 0.9873                                                                                 & \textbf{0.8892}                                                                                 & 0.9082                                                                                            & \textbf{0.8481}                                                                                                 & 0.9778                                                                                              & 0.9937                                                                                                            & \textbf{0.8323}                                                                                  & 0.6677                                                                                                   & \textbf{0.6804}                                                                                               \\ \hline
           & \cellcolor[HTML]{EFEFEF}\begin{tabular}[c]{@{}l@{}}Pointing to\\ Something with\\ Finger\end{tabular}   & \cellcolor[HTML]{EFEFEF}\begin{tabular}[c]{@{}l@{}}Taking a\\ Selfie\end{tabular}      & \cellcolor[HTML]{EFEFEF}\begin{tabular}[c]{@{}l@{}}Check Time\\ (from watch)\end{tabular}       & \cellcolor[HTML]{EFEFEF}\textbf{\textit{\begin{tabular}[c]{@{}l@{}}Rub Two\\ Hands \\ Together\end{tabular}}}       & \cellcolor[HTML]{EFEFEF}Nod Head/Bow                                                                            & \cellcolor[HTML]{EFEFEF}Shake Head                                                                  & \cellcolor[HTML]{EFEFEF}\textit{\textbf{Wipe Face}}                                                               & \cellcolor[HTML]{EFEFEF}Salute                                                          & \cellcolor[HTML]{EFEFEF}\begin{tabular}[c]{@{}l@{}}Put the \\ Palms Together\end{tabular}                & \cellcolor[HTML]{EFEFEF}\begin{tabular}[c]{@{}l@{}}Cross Hand \\ in Front\end{tabular}                        \\ \hline
Baseline   & 0.8540                                                                                                  & 0.8111                                                                                 & 0.8677                                                                                          & \textbf{0.5962}                                                                                            & 0.9323                                                                                                          & 0.8582                                                                                              & \textbf{0.6196}                                                                                                   & 0.9073                                                                                  & 0.8519                                                                                                   & 0.9067                                                                                                        \\ \hline
MS-Res-TCN & 0.8476                                                                                                  & 0.8386                                                                                 & 0.8544                                                                                          & \textbf{0.6741}                                                                                            & 0.9367                                                                                                          & 0.8797                                                                                              & \textbf{0.7500}                                                                                                   & 0.9367                                                                                  & 0.8829                                                                                                   & 0.9391                                                                                                        \\ \hline
           & \cellcolor[HTML]{EFEFEF}\textit{\textbf{Sneeze/Cough}}                                                  & \cellcolor[HTML]{EFEFEF}Staggering                                                     & \cellcolor[HTML]{EFEFEF}Falling                                                                 & \cellcolor[HTML]{EFEFEF}Touch Head                                                                & \cellcolor[HTML]{EFEFEF}\begin{tabular}[c]{@{}l@{}}Touch \\ Chest\end{tabular}                                  & \cellcolor[HTML]{EFEFEF}Touch Back                                                                  & \cellcolor[HTML]{EFEFEF}\textit{\textbf{Touch Neck}}                                                              & \cellcolor[HTML]{EFEFEF}Nausea                                                          & \cellcolor[HTML]{EFEFEF}\begin{tabular}[c]{@{}l@{}}Use a Fan / \\ Feeling Warm\end{tabular}              & \cellcolor[HTML]{EFEFEF}\textbf{\textit{\begin{tabular}[c]{@{}l@{}}Punching /\\ Slapping\end{tabular}}}                         \\ \hline
Baseline   & \textbf{0.7158}                                                                                         & 0.8899                                                                                 & 0.9867                                                                                          & 0.7316                                                                                            & 0.8085                                                                                                          & 0.6949                                                                                              & \textbf{0.6006}                                                                                                   & 0.8500                                                                                  & 0.7750                                                                                                   & \textbf{0.7905}                                                                                                        \\ \hline
MS-Res-TCN & \textbf{0.8291}                                                                                         & 0.9146                                                                                 & 0.9873                                                                                          & 0.7500                                                                                            & 0.8006                                                                                                          & 0.7184                                                                                              & \textbf{0.7595}                                                                                                   & 0.9082                                                                                  & 0.7975                                                                                                  & \textbf{0.8590}                                                                                                        \\ \hline
           & \cellcolor[HTML]{EFEFEF}\begin{tabular}[c]{@{}l@{}}Kicking\\ Other\\ Person\end{tabular}                & \cellcolor[HTML]{EFEFEF}\begin{tabular}[c]{@{}l@{}}Pushing \\Other\\ Person\end{tabular} & \cellcolor[HTML]{EFEFEF}\begin{tabular}[c]{@{}l@{}}Pat on Back \\ of Other Person\end{tabular}  & \cellcolor[HTML]{EFEFEF}\begin{tabular}[c]{@{}l@{}}Point Finger at\\ Other Person\end{tabular}    & \cellcolor[HTML]{EFEFEF}\begin{tabular}[c]{@{}l@{}}Hugging\\ Other\\ Person\end{tabular}                        & \cellcolor[HTML]{EFEFEF}\textbf{\textit{\begin{tabular}[c]{@{}l@{}}Giving \\Something to\\Other Person\end{tabular}}} & \cellcolor[HTML]{EFEFEF}\textit{\textbf{\begin{tabular}[c]{@{}l@{}}Touch Other\\ Person's\\ Pocket\end{tabular}}} & \cellcolor[HTML]{EFEFEF}\textit{\begin{tabular}[c]{@{}l@{}}Hand\\ Shaking\end{tabular}} & \cellcolor[HTML]{EFEFEF}\textit{\begin{tabular}[c]{@{}l@{}}Walking\\ Towards \\ Each Other\end{tabular}} & \cellcolor[HTML]{EFEFEF}\textit{\begin{tabular}[c]{@{}l@{}}Walking\\ Apart\\ from \\ Each Other\end{tabular}} \\ \hline
Baseline   & 0.8673                                                                                                  & 0.9300                                                                                 & 0.9007                                                                                          & 0.9166                                                                                            & 0.9651                                                                                                          & \textbf{0.8424}                                                                                              & \textbf{0.7997}                                                                                                   & 0.9280                                                                                  & 0.9492                                                                                                   & 0.9455                                                                                                        \\ \hline
MS-Res-TCN & 0.9010                                                                                                  & 0.9340                                                                                 & 0.8651                                                                                          & 0.9040                                                                                            & 0.9732                                                                                                          & \textbf{0.9112}                                                                                              & \textbf{0.9148}                                                                                                   & 0.9276                                                                                  & 0.9547                                                                                                   & 0.9677                                                                                                        \\ \hline
\end{tabular}%
}
\end{table*}
\subsection{Interpretable Model Diagnosis}
\label{section:exp_interp}
\begin{figure}[b]
\begin{center}
 \includegraphics[width=7cm,height=7cm]{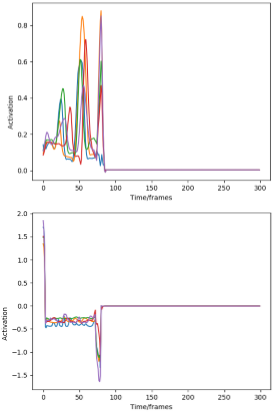}
\end{center}
\label{fig:curve}
 \caption{The figures display filter response magnitudes plotted over time. \textbf{Top}: Response plots of 5 samples with action class \textit{Jump Up}. The filters show  \textbf{Bottom}: Response plots of 5 samples with action class \textit{Reading}. Lack of filter responses for \textit{Reading} indicates that Res-TCN has failed to learn motion primitives to discriminate this action as opposed to well-converged action class of \textit{Jump Up}. }
\end{figure}

Feature Map Decoder takes an arbitrary hidden representation $X_l$ from Res-TCN and outputs a sequence of moving skeletons. The visualized results are shown in Figures \ref{fig:magnify}, \ref{fig:view} and \ref{fig:hand}. Through the proposed diagnosis, we identified three characteristic learning patterns of Res-TCN:

\begin{itemize}[leftmargin=*,nosep]
\item 
The model learns a set of discriminative joints for actions and the movement of these joints are amplified in deeper layers. For example, as visualized on the left of Figure \ref{fig:magnify} the motion of joints related to legs, head and neck are greatly magnified in deeper layers for action class \textit{jump up}. This makes intuitive sense that the vertical movement of the head is a rare occurrence in the training data compared to other human actions in the dataset. Similar learning pattern is observed for action class \textit{cheer up} where vertical displacement of the hand joints is significantly amplified as the layers get deeper.
\item 
In the case of NTU, the actions are recorded from multiple view angles ($-45^{\circ}$, $0^{\circ}$ and $+45^{\circ}$). Compared to other models \cite{Authors1,Authors2,Authors5} that manually normalize and rotate the skeleton to a frontal view, we do not perform \textit{any} scale or rotational normalization. The sequences in Figure \ref{fig:view} depict the same action recorded from different view angles. Interestingly, the decoded skeleton sequences in Figure \ref{fig:view} show that the model has learned to rotate the skeleton sequence implicitly to a more discriminative viewpoint (frontal view). Moreover, the decoded sequences provide strong evidence that the model has achieved view-point invariance. Despite given inputs from different view points, the model has learned to produce a common representation for both inputs as evidenced by the similarity of the decoded skeletons. 
\item 
Res-TCN fails to learn a representation for a detailed fine-grained motion of the hands. The decoded sequences for action classes such as \textit{Reading} and \textit{Writing} clearly highlight that the model failed to learn a discriminative hidden representation. As seen in Figure \ref{fig:hand}, the decoded skeletons for \textit{Reading} and \textit{Writing} show significant resemblance. Moreover, as depicted in Figure \ref{fig:curve}, unlike well-discriminative action classes such as \textit{Jump Up}, response magnitudes of filters show no characteristic signature or no response at all for cases such as \textit{Reading}.
\end{itemize}

\subsection{Targeted Refinement: Multi-stream Residual Temporal Convolutional Networks}
The diagnosis analysis revealed Res-TCN's weakness that it can not recognize fine-grained hand movements. We aimed to pin point this shortcoming of our model through the targeted refinement technique discussed in Section \ref{refinement}. Table 1 shows that the diagnosis based refinement approach (MS-Res-TCN) improves the classification performance of the baseline Res-TCN model and achieves the state-of-the-art on NTU benchmark. We also provide a detailed performance comparison between the models on the NTU benchmark in Table 2. The highlighted cells in the table indicate where the MS-Res-TCN has improved upon the baseline Res-TCN by a healthy margin. The highlighted action classes share a common trait that the actions involve nuanced hand motion which validates the effectiveness of the proposed targeted refinement approach.

\section{Conclusion}
Through the 'train-diagnose-and-fix' paradigm, this work proposes an interpretable, intuitive yet an effective approach for learning powerful models for fine-grained action recognition. The proposed Feature Map Decoder provides an useful model interpretation and diagnosis tool which exposes key learning traits of the model. Through the analysis, we revealed that the model has learned to implicitly rotate the input sequence to a more discriminative view and gradually emphasize the characteristic spatio-temporal patterns of the input. We also demonstrate that our targeted refinement technique is an effective and a systematic procedure to encode newly gained intuition into our models. MS-Res-TCN, the resulting model after one iteration of the 'train-diagnose-and-fix' cycle, validates that the proposed approach is not only intuitive but also potent in learning discriminative representations for fine-grained skeleton based action recognition tasks.


{\small
\bibliographystyle{ieee}
\bibliography{biblio}
}

\end{document}